# Learning Sparse Visual Representations with Leaky Capped Norm Regularizers

**Jianqiao Wangni**[*]
Tsinghua University
Beijing, China
zjnqha@gmail.com

**Dahua Lin**
The Chinese University of Hong Kong
Hong Kong, China
dhlin@ie.cuhk.edu.hk

## Abstract

Sparsity inducing regularization is an important part for learning over-complete visual representations. Despite the popularity of $\ell_1$ regularization, in this paper, we investigate the usage of non-convex regularizations in this problem. Our contribution consists of three parts. First, we propose the leaky capped norm regularization (LCNR), which allows model weights below a certain threshold to be regularized more strongly as opposed to those above, therefore imposes strong sparsity and only introduces controllable estimation bias. We propose a majorization-minimization algorithm to optimize the joint objective function. Second, our study over monocular 3D shape recovery and neural networks with LCNR outperforms $\ell_1$ and other non-convex regularizations, achieving state-of-the-art performance and faster convergence. Third, we prove a theoretical global convergence speed on the 3D recovery problem. To the best of our knowledge, this is the first convergence analysis of the 3D recovery problem.

## Introduction

The sparse models have been widely applied in machine learning and computer vision tasks. To encourage sparsity, the $\ell_1$ norm is widely adopted as a regularizer, which can produce reasonable results in various cases. For an example, to analyse the 3D shape of an object, from a monocular 2D camera, one can build a dictionary of 3D shape bases from datasets, and inference the sparse combination weights of the bases (Zhou et al. 2016). Yet, a number of studies suggested that it is not always the optimal choice, *non-convex* regularizers can often lead to better balance between sparsity and accuracy in practice. Particularly, a number of non-convex regularizers have been explored in previous work, such as $\ell_p$ norm (Frank et al. 1993), MCP (Zhang 2010a), SCAD (Fan and Li 2011), Logarithm (Friedman 2012), and capped norm (Zhang 2010b)(Zhang and others 2013). However, the studied models are mostly restricted to be convex ones, therefore lack suitability to larger application scope. This is reasonable since, first, the convex problem with non-convex regularizers are already difficult to analyse, the global optima can only be obtained with additional statistical assumptions. The non-convex problem, even without regularizers, is much more difficult to optimize, the solver for non-convex will converges much slower comparing to when it applies to convex ones.

In this work, we will concentrate on the learning better sparse models, with considerations on the running time limitation, which commonly exists, especially in testing stage. For an example, the aforementioned 3D shape recovery models have achieved good enough accuracy, but they converge too slow for real-time applications. Although directly enlarging the $\ell_1$ regularizer will force more small weights to be zeros within limited time, but this will certainly compromise the accuracy since the larger and effective weights are also penalized strongly. In this work, we aim to move beyond the limitations of existing literatures and develop a novel regularizer that can achieve high sparsity and high accuracy in learning sparse representations (like other successful applications based on non-convex regularizers), The key to achieve this goal is to force more smaller weights to zeros by a stronger regularization, but not so far as to leading larger weights deviating from the ground truth. Following this idea, we propose a novel non-convex regularizer, called *Leaky Capped Norm*, which is a generalization of the capped norms, but enjoys nice mathematical properties and demonstrates higher performance in experiments. We also derive a doubly majorization-minimization algorithm, which is suitable for both convex and nonconvex loss functions. We also take the 3D shape recovery problem for detailed analysis. By relaxing the objective to a convex one, we can apply an ADMM solver and get a theoretical global convergence. We also conducted experiments on sparse neural networks to show the efficiency in compressing kernel weights.

## Leaky Capped $\ell_1$ Norm

A standard sparse model consists of a loss function $L$ (which we assume to be differentiable but possibly nonconvex) and a regularization part $H$ on the weights $c$

$$\min_c \frac{1}{N} \sum_n L(x_n, c) + H(c). \quad (1)$$

where $\{x_n\}$ are data points. A popular choice for $H$ is the $\ell_1$ norm, due to its convinience for optimization and that it is good surrogate for $\ell_0$ norm. There are also some literatures suggesting that the $\ell_1$ norm is in many case inferior to non-convex norms for inducing sparsity. To motivate our new regularizer, we begin with a discussion on the representative capped $\ell_1$ norm (Zhang 2010b):

$$H(c) = \alpha \sum_i \min(|c_i|, \tau), \quad \text{where} \quad \alpha, \tau > 0. \quad (2)$$

---
[*]This work was done when the first author was a visiting research assistant at CUHK.

As shown in Figure 1, this formulation only regularizes those weights that are below a certain threshold $\tau$. For those beyond this value, they can grow arbitrarily without experiencing penalties. It is noteworthy that it is a generalization of $\ell_1$ norm. Particularly, it becomes $\ell_1$ as $\tau \to \infty$. Generally, with a finite $\tau$, it approximates $\ell_0$ better than $\ell_1$, and therefore leads to higher sparsity in some real-world applications. The key feature of capped norm is the lack of penalty for weights whose magnitudes are greater than $\tau$, this feature, however, may become a drawback under some circumstances. For examples, in our empirical study on 3D shape recovery, which is introduced later, the capped norm leads to unstable optimization procedures and substantially worse solutions, by the multi-stage algorithm introduced later. Another difficulty shared by both $\ell_1$ and capped $\ell_1$ norms is the parameters selection – it is often very tricky to find a good value of $\tau$ and $\alpha$ that well balances between accuracy and sparsity. From an optimization perspective, the two parameters are both critical to the most common proximal operator, as

$$\arg\min_x \frac{1}{2}||x - c||^2 + H(x) = \text{sign}(c)\max(|c| - \mathbb{I}(c < \tau)\alpha, 0),$$

where $\mathbb{I}(\cdot)$ is the indicator function that equals to one for truth and zero for false. As implied by theoretical analysis and shown in our experiments, larger values of $\alpha$ and $\tau$ would yield sparser solutions, but at the same time damage the larger weights and consequently compromise the accuracy. With the analysis above in mind, we propose the *Leaky Capped $\ell_1$ Norm Regularizer (LCNR)* as an improved variant:

$$H(c) = \alpha \sum_i \min(|c_i|, \tau) + \beta \sum_i \max(|c_i|, \tau), \quad (3)$$

where $0 < \beta < \alpha$. As shown in Figure 1, the function of LCNR is piecewise linear and thus is differentiable (except at a few points). The key difference between the proposed formulation and the standard $\ell_1$ norm and the capped norm is that large weights (*i.e.* those greater than $\tau$) are still penalized *positively* but *less heavily*. To be more specific, this generalizes the capped $\ell_1$ norm, with an additional coefficient $\beta$ to control how much those large weights are regularized. In particular, when $\beta = 0$, it reduces to the regular capped norm.

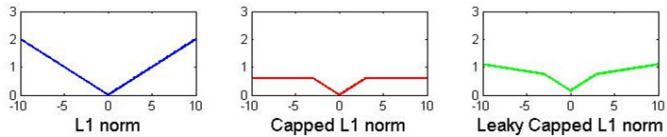

Figure 1: Geometric view of three kinds of norms.

A representative optimization algorithm for solving convex problems with nonconvex regularizer is the multi-stage algorithm (Zhang 2010b), it is guaranteed to converge since it is essentially a special case of the majorization-minimization (MM) algorithm (Hunter and Lange 2004). To be more specific, at the $(l + 1)$-th stage, the surrogate objective is

$$H^{l+1}(c) = |c^T| \cdot \lambda^l, \quad \text{where} \quad \lambda_i^l = \alpha\mathbb{I}(|c_i^l| \leq \tau) + \beta\mathbb{I}(|c_i^l| > \tau),$$

where $c^l$ is the optimal $c$ at the $l$-th stage, but the surrogate loss remains the same as original one. We set the initialization parameter $\lambda_i^0 = \beta$, so as to begin with a light regularization for all weights. Based on the algorithm, one can see that the parameter selection is simpler - one can first search $\beta$ for accuracy and then $\alpha$ for sparsity.

## Monocular 3D Shape Recovery

Now we proceed to demonstrating the usage of the proposed regularizer and algorithms for solving a computer vision problem, the 3D shape analysis of visual objects. Driven by the demands arising from real-world applications, 3D shape recovery from 2D images has received increasing attention from the computer vision community in recent years. A key challenge of this task lies in the fact that different 3D shapes can be projected to the same 2D image, resulting in an ill-posed problem. A commonly adopted approach is to restrict the recovered 3D shape to be a linear combination of a set predefined shape bases. A representative model in early ages along this line is the Active Shape Model (ASM)(Cootes et al. 1995), which relies on a dense combination for shape representation. Later advances, like (Hejrati and Ramanan 2012)(Zia et al. 2013)(Wang et al. 2014))(Xiang and Savarese 2012)(Ramakrishna, Kanade, and Sheikh 2012)(Lin et al. 2014)however, shows that sparse representation is more effective in real-world applications, often demonstrating stronger generalization performance and higher robustness against adverse conditions.

The 3D shape, i.e. the 3D locations of $p$ landmarks are stacked as $S \in \mathbb{R}^{3 \times p}$, and its corresponding 2D shape is $Y \in \mathbb{R}^{2 \times p}$. We denote the projection matrix as $\Pi$. The Active Shape Models (ASM) (Cootes et al. 1995) proposed training a group of shape bases $\{B_i\}_{i=1}^D$ from data by methods like principal component analysis, denoting $\mathcal{D} = \{1, \cdots, D\}$ as a set of subscripts and $c$ as weights of each basis. The 3D-2D shape relation is characterized as

$$S = \sum_{i \in \mathcal{D}} c_i B_i, \quad Y = \Pi S,$$

$$\text{where} \quad \Pi = \begin{pmatrix} \omega & 0 & 0 \\ 0 & \omega & 0 \end{pmatrix}, c \in \mathbb{R}^D, B_i \in \mathbb{R}^{3 \times p} \quad (4)$$

here $\omega$ is a parameter depending on physical factors like focal length and view depth. In the test phase, the 2D shape $Y$ is annotated by regular visual detectors, since the shape bases $B$ are most likely predefined in a different camera setting other than the test setting, the unknown factors including combination weights $c$, a relative rotation parameter $R$ and a translation parameter $T$ should all be considered in the projection model,

$$Y = \Pi(R \sum_{i \in \mathcal{D}} c_i B_i + T), \text{ where } R \in SO(3), T \in \mathbb{R}^3 \quad (5)$$

where $I_3 \in \mathbb{R}^{3 \times 3}$ is an identity matrix and $SO(3) = \{R \in \mathbb{R}^{3 \times 3} \mid R^T R = I_3, \det(R) = 1\}$. In addition, (Zhou et al. 2016) proposed distributing individual rotation matrix $R_i \in \mathbb{R}^{3 \times 3}$ to each basis, then the 3D and 2D shapes are represented as $= \sum_{i \in \mathcal{D}} c_i R_i B_i$. They substitute the bilinear term composed of $\Pi$ and $R$ by uniform variables $\{M_i \in \mathbb{R}^{2 \times 3}\}_{i \in \mathcal{D}}$, that $M_i = c_i \Pi R_i$ which implicitly take rotation and projection factors into account. Denoting $M$ as a 3 dimensional tensor stacking $\{M_i\}_{i=1}^{D}$, we rewrite the objective as

$$\min_{R_i \in \Omega(c_i), c} F(M, c) = \frac{1}{2} ||Y - \sum_{i \in \mathcal{D}} M_i B_i||_F^2 + H(c), \quad (6)$$

$$\text{where } \Omega(c_i) = \{M_i \in \mathbb{R}^{2 \times 3} | M_i^T M_i = c_i^2 I_2\},$$

and $H(c)$ is the regularization and $I_2 \in \mathbb{R}^{2 \times 2}$ is an identity matrix. In our model of 3D shape recovery, we adopt the linear formulation as Eq.(6) since it is the best baseline up until now and has some attractive properties. Our objective function is same with Eq.(6), $H(c)$ is set to be LCNR as Eq.(3). The objective with a stage-wise surrogate regularizer is

$$F^{l+1}(M, c) = \frac{1}{2} ||Y - \sum_{i \in \mathcal{D}} M_i B_i||_F^2 + \sum_{i \in \mathcal{D}} \lambda_i^l |c_i|, \quad (7)$$

in the $(l+1)$-th stage, where $M_i \in \Omega(c_i)$. In this optimizer, the regularization part of $F(R, c)$ is convex in each stage, however, the surrogate function is still non-convex since there is an orthogonality constraint on each $M_i$. To build the upper bounding function, we make a convex relaxation on the constraints. As the spectral-norm ball $conv(\Omega(c_i)) = \{X \in \mathbb{R}^{2 \times 3} \mid ||X||_2 \leq c_i\}$, is the tightest convex hull of the Stiefel manifold $\Omega(c_i)$ (Zhou et al. 2016) [(Journée et al. 2010), Section 3.4], where $|| \cdot ||_2$ represents the spectral norm, which is its largest singular value. Finally, by relaxing the domain $\Omega$ to $conv(\Omega)$, and the recalibration rule of $\lambda_i$ is transferred to $\lambda_i^l = \alpha \mathbb{I}(||\hat{M}_i^l||_2 \leq \tau) + \beta \mathbb{I}(||\hat{M}_i^l||_2 > \tau)$, where $\hat{M}_i^l$ is the optimal $M_i$ in the $l$-th stage.

Since the function is convexified, We employ the alternate direction method of multiplier (ADMM) (Boyd et al. 2011)(Zhou et al. 2016) algorithm to attain high-precision solutions. We introduce a tensor $V$ as a copy of $M$, $U$ as a dual tensor variable and $\mu$ as a stepsize parameter, then rewrite Eq.(7) in its augmented Lagrangian formulation

$$F_\mu^{l+1}(M, V, U) = \frac{1}{2} ||Y - \sum_{i \in \mathcal{D}} V_i B_i||_F^2 + \sum_{i \in \mathcal{D}} \lambda_i^l ||M_i||_2$$
$$+ \sum_{i \in \mathcal{D}} U_i^T (M_i - V_i) + \frac{\mu}{2} \sum_{i \in \mathcal{D}} ||M_i - V_i||^2 \quad (8)$$

Then the ADMM procedure is applied to solve the subproblem. After the convergence, the multi-stage solver will update the surrogate functions. We denote the inner-iteration superscript as $t$. Then $M^{t+1}$ is update based on the proximal operator on spectral norms [(Parikh, Boyd, and others 2014), Section 6.7.2],

$$prox_\lambda(V_i') = P \operatorname{diag}[\sigma - \lambda_i' \mathcal{P}_1(\sigma/\lambda_i')] Q^T, \quad (9)$$

where $V_i' = V_i^t - U_i^t/\mu$ and $\lambda_i' = \lambda_i^l/\mu$. Denoting the solution as $prox_{\lambda_i'}(V_i')$, $V_i' = P \operatorname{diag}(\sigma) Q^T$ is the singular value decomposition of $V_i'$, and $\mathcal{P}_1(\cdot)$ is the Euclidean projection onto the $\ell_1$ norm ball. The update on $V$ and $U$ have closed form solutions,

$$V_i^{t+1} = (Y B_i^T + \mu M_i^{t+1} + U_i^{t+1})(B_i B_i^T + \mu I_3)^{-1}, \quad (10)$$
$$U_i^{t+1} = U_i^t + \mu(M_i^t - V_i^t). \quad (11)$$

The convergence property of this algorithm is well studied in (Boyd et al. 2011), additionally, we adopt an adaptive policy for stepsize $\mu$ as suggested by [(Boyd et al. 2011), Section 3].

### Theoretical Analysis

Generally speaking, the convergence of optimization over non-convex function is typical hard to prove without a decreasing step size. When combined with nonconvex regularizers, even convex problems need highly specialized proofs, for each problem. One may see the hardness of by referring to literatures, on the theoretical analysis of capped norm in different applications, *e.g.*, multi-task feature learning (Gong, Ye, and Zhang 2012)(Tang, Nie, and Jain 2016), matrix completion (Gao et al. 2015) and (Jiang, Nie, and Huang 2015)(Sun, Xiang, and Ye 2013)(Han and Zhang 2016)(Zhang 2010b)(Zhang and others 2013). The proof is harder for our problem since the loss function is already non-convex. In this section, we theoretically prove that with a high probability, the recovery error of our multi-stage algorithm decreases at nearly exponential speed against stages. We assume that the ground truth of the 2D shape $\bar{Y} \in \mathbb{R}^{2 \times p}$ is expressed as a projection of the combined deformation of 3D shape bases, as $\bar{Y} = \sum_{i \in \mathcal{D}} \bar{M}_i B_i$, where $\bar{M}_i$ is the ground truth of deformation matrix $M_i$, for $0 \leq i \leq D$. The observation model is $Y = \bar{Y} + \delta$, where $\delta \in \mathbb{R}^{2 \times p}$ is a Gaussian noise, i.e. $\delta_{jk} \sim \mathcal{N}(0, \sigma^2)$. For notational simplicity, we also set $\gamma = \alpha + \beta$.

**Assumption 1** *For any matrix $M_i \in \mathbb{R}^{n \times m}$, we assume that there exist a constant $\kappa_i$ that*

$$\kappa_i = \min_{M_i \in \mathcal{R}(s)} ||Z - M_i B_i||_F / ||M_i||_\star > 0, \quad (12)$$

*where the restricted set $\mathcal{R}(s)$ is defined as $\mathcal{R}(s) = \{X \in \mathbb{R}^{n \times m} \mid rank(X) \leq s\}$.*

**Remark 1** *This is the widely used eigenvalue assumption which can be found in (Lounici et al. 2009).*

**Theorem 2** *Following the common setting of sparse dictionary learning, we assume that each basis $B_i$ are normalized by row $\ell_2$ norm that $\sum_k B_{irk}^2 = \phi$ for all $i \in \mathcal{D}, 1 \leq r \leq 3$, where $\phi$ is an constant. For the optimal matrix $\hat{M}_i \in \mathbb{R}^{2 \times 3}$ in any stage, if we set $\alpha, \beta$ as $(\alpha + \beta) \geq \phi \sqrt{3 + e}/2$, then it*

holds
$$\frac{1}{2}||\bar{Y} - \sum_{i \in \mathcal{D}} \hat{M}_i B_i||_F^2 \leq \frac{1}{2}||\bar{Y} - \sum_{i \in \mathcal{D}} M_i B_i||_F^2 + \sum_{i \in \mathcal{D}} (4\gamma + \lambda_i^l)||M_i - \hat{M}_i||_2, \quad (13)$$

with the probability of at least $1 - 2D \exp(-\frac{1}{2}(e - 3\ln(1 + e/3)))$, where $e$ is a positive scalar.

**Proof.** Recalling that $Y = \bar{Y} + \delta$ and the property of optimal point $\hat{M}_i$, then we have

$$\frac{1}{2}||\bar{Y} - \sum_{i \in \mathcal{D}} \hat{M}_i B_i||_F^2 \leq \frac{1}{2}||\bar{Y} - \sum_{i \in \mathcal{D}} M_i B_i||_F^2 \quad (14)$$

$$+ \sum_{i \in \mathcal{D}} \lambda_i^l ||M_i - \hat{M}_i||_2 + \sum_{i \in \mathcal{D}} \text{tr}[(M_i - \hat{M}_i) B_i \delta^T], \quad (15)$$

where we use the triangular inequality of spectral norm

$$||M_i||_2 - ||\hat{M}_i||_2 \leq ||M_i - \hat{M}_i||_2. \quad (16)$$

We first establish the upper bound of $\text{tr}[(\hat{M}_i - M_i) B_i \delta^T]$. We denote a set of random events $\{\mathcal{A}_{ij}\}$ and define a set of random variables $\{v_{ijr}\}$ as

$$\mathcal{A}_{ij} = \{||B_i^T \delta_j||_2 \leq \phi\gamma\}, \quad v_{ijr} = \frac{1}{\phi}\sum_{k=1}^{p} B_{irk}\delta_{jk}, \quad (17)$$

where $B_{irk}$ is the element in the $r$-th row and $k$-th column of $B_i$. Since $B_i$ is normalized, $v_{ijr}$ are i.i.d. Gaussian variables following $\mathcal{N}(0, 1)$. Then we can verify that $\sum_{r=1}^{3} v_{ijr}^2$ is a chi-squared random variable with $d = 3$ degree of freedom. By choosing $\lambda$ according to Theorem 2, we have

$$Pr(\frac{1}{2}||B_i \delta_j^T||_2 > \gamma) = Pr(\sum_{r=1}^{3}(\sum_{k=1}^{p} B_{irk}\delta_{jk})^2 > 4\gamma^2)$$

$$\leq Pr(\sum_{r=1}^{3} v_{ijr}^2 > 3 + e) \leq \exp(-\frac{1}{2}\theta(e)), \quad (18)$$

where $\theta(e) = e - 3\ln(1 + e/3)$ and the second inequality is due to the chi-squared distribution (Chen, Zhou, and Ye 2011). Denoting $\mathcal{A} = \bigcap_{i=1}^{D} \bigcap_{j=1}^{2} \mathcal{A}_{ij}$, we also denote $\mathcal{A}^c$ as its complementary set and $|\mathcal{A}|$ as its cardinality, then

$$Pr(\mathcal{A}) = 1 - \bigcup_{i=1}^{D} \bigcup_{j=1}^{2} \mathcal{A}_{ij}^c \geq 1 - 2D \exp(-\frac{1}{2}\theta(e)). \quad (19)$$

Denoting $M_{ir}$ as the $r$-th row of $M_i$, we can derive an upper bound on $\text{tr}[(\hat{M}_i - M_i) B_i \delta^T]$ under the event $\mathcal{A}$,

$$\text{tr}[(M_i - \hat{M}_i) B_i \delta^T] = \sum_{r=1}^{2}\sum_{j=1}^{2} (M_{ir} - \hat{M}_{ir})^T B_i \delta_j^T$$

$$\leq \sum_{r=1}^{2}\sum_{j=1}^{2} ||M_{ir} - \hat{M}_{ir}||_2 ||B_i \delta_j^T||_2 \leq 4\gamma ||M_i - \hat{M}_i||_2, \quad (20)$$

where we apply the Cauchy-Schwarz inequality and the relation between Frobenius norm and spectral norm. By substituting this back to Eq.(14), we get the proof.

**Theorem 3** *Let $\hat{M}_i^{l+1}$ be the optimal solution at the $(l+1)$-th stage, and $\hat{M}_i^l$ be the one at the $l$-th stage accordingly. We define $W_i^l = \bar{M}_i - \hat{M}_i^l$, and a function $\mathcal{L}$ on set $\mathcal{S} \subseteq \mathcal{D}$. If we choose $\alpha, \beta$ as in Theorem 1 and choose $\tau$ as $\tau > (\alpha + \beta)/\kappa^2$, the following inequality stands*

$$\mathcal{L}_{l+1}(\mathcal{D}) \leq a^l \mathcal{L}_0(\mathcal{D}) + \frac{b}{1-a}, \quad \text{where} \quad \mathcal{L}_l(\mathcal{S}) = \sqrt{\sum_{i \in \mathcal{S}} ||W_i^l||_2^2},$$

*with probability of at least $1 - 2D \exp(-\frac{1}{2}(e - 3\ln(1 + e/3)))$, where $a = (\alpha + \beta)/(\kappa^2 \tau)$, and $b = 5(\alpha + \beta)\sqrt{D}/(\kappa^2 \tau)$.*

**Proof.** We apply Theorem 2 in stage $(l+1)$ and substitute $M$ by its ground truth $\bar{M}$, then get

$$\frac{1}{2}||\bar{Y} - \sum_{i \in \mathcal{D}} \hat{M}_i^{l+1} B_i||_F^2 \leq \sum_{i \in \mathcal{D}} (4\gamma + \lambda_i^l) ||W_i^{l+1}||_2, \quad (21)$$

where we use $\bar{Y} = \sum_{i \in \mathcal{D}} \bar{M}_i B_i$. We define a set $\mathcal{G} = \{i \in \mathcal{D} \mid ||\hat{M}_i^l||_2 \leq \tau\}$ to separate the weights, and

$$\alpha_i^l = \alpha \mathbb{I}(i \in \mathcal{G}), \quad \beta_i^l = \beta \mathbb{I}(i \in \mathcal{G}^c), \quad (22)$$

then there is $\lambda_i^l = \alpha_i^l + \beta_i^l$. Then we establish a bound by Cauchy-Schwarz inequality,

$$\sum_{i \in \mathcal{D}}(\lambda_i^l + 4\gamma)||W_i^{l+1}||_2 = \sum_{i \in \mathcal{D}}(\alpha_i^l + \beta_i^l + 4\gamma)||W_i^{l+1}||_2$$

$$\leq (4\gamma\sqrt{D} + \alpha\sqrt{|\mathcal{G}|} + \beta\sqrt{|\mathcal{G}^c|})\mathcal{L}_{l+1}(\mathcal{D}). \quad (23)$$

To further bound this term, we first denote that

$$\mathcal{E} = \{i \in \mathcal{D} \mid ||\bar{M}_i||_2 \neq 0\}, \mathcal{F} = \{i \in \mathcal{D} \mid ||\bar{M}_i||_2 \leq 2\tau\}, (24)$$

by the rule of set operation and the definition of $\mathcal{G}$ and $\mathcal{F}$,

$$|\mathcal{G}| = |\mathcal{G} \cap \mathcal{F}| + |\mathcal{G} \cap \mathcal{F}^c|, \quad \text{where} \quad |\mathcal{G} \cap \mathcal{F}| \leq |\mathcal{F}|, \quad (25)$$

$$\tau^2 |\mathcal{G} \cap \mathcal{F}^c| \leq \sum_{i \in \mathcal{G} \cap \mathcal{F}^c} ||\bar{M}_i - \hat{M}_i^l||_2^2 \leq \mathcal{L}_l^2(\mathcal{G} \cap \mathcal{F}^c); \quad (26)$$

by the inequality $||\bar{M}_i - \hat{M}_i^l||_2 \geq ||\bar{M}_i||_2 - ||\hat{M}_i^l||_2 \geq \tau$. Substituting them back to Eq.(23), we get

$$\sum_{i \in \mathcal{D}} \alpha_i^l ||W_i^{l+1}||_2 \leq \alpha\sqrt{|\mathcal{F}| + \mathcal{L}_l^2(\mathcal{F}^c)/\tau^2} \mathcal{L}_{l+1}(\mathcal{D}) \quad (27)$$

$$\leq (\alpha\sqrt{|\mathcal{F}|} + \frac{\alpha}{\tau}\mathcal{L}_l(\mathcal{F}^c))\mathcal{L}_{l+1}(\mathcal{D}), \quad (28)$$

where in the last inequality we use $\sqrt{a^2 + b^2} \leq a + b$ for $a, b \geq 0$. A similar result holds for another part of Eq.(23) as

$$\sum_{i \in \mathcal{D}} \beta_i^l ||W_i^{l+1}||_2 \leq (\beta\sqrt{|\mathcal{E}|} + \frac{\beta}{\tau}\mathcal{L}_l(\mathcal{E}^c))\mathcal{L}_{l+1}(\mathcal{D}). \quad (29)$$

Substituting them back to Eq.(21), there is

$$\frac{1}{2}||\bar{Y} - \sum_{i \in \mathcal{D}} \hat{M}_i^{l+1} B_i||_F^2 \leq \sum_{i \in \mathcal{D}} (4\gamma + \lambda_i^l)||W_i^{l+1}||_2$$

$$\leq \gamma(4\sqrt{D} + \sqrt{\max(|\mathcal{E}|, |\mathcal{F}|)} + \frac{1}{\tau}\mathcal{L}_l(\mathcal{D}))\mathcal{L}_{l+1}(\mathcal{D}), \quad (30)$$

Recalling Assumption 1 and substituting $Z = \bar{M}_i B_i$, then

$$\kappa_i^2 ||W_i^{l+1}||_2^2 \leq \kappa_i^2 ||W_i^{l+1}||_\star^2 \leq \frac{1}{2}||W_i^{l+1} B_i||_F^2, \quad (31)$$

where we use $||X||_F \leq ||X||_2$. Denoting $\kappa = \min_i \kappa_i$, then

$$2\kappa^2 \sum_{i\in\mathcal{D}} ||W_i^{l+1}||_2^2 \leq \sum_{i\in\mathcal{D}} ||W_i^{l+1} B_i||_F^2 \leq ||\sum_{i\in\mathcal{D}} W_i^{l+1} B_i||_F^2.$$

Substituting this to Eq.(21) and combining for $i \in \mathcal{D}$, there is

$$\kappa^2 \mathcal{L}_{l+1}^2(\mathcal{D}) \leq (\alpha+\beta)(5\sqrt{D} + \frac{1}{\tau}\mathcal{L}_l(\mathcal{D}))\mathcal{L}_{l+1}(\mathcal{D}). \quad (32)$$

where we apply $\max(|\mathcal{E}|, |\mathcal{F}|) \leq D$. Recalling the definition of $a$ and $b$, we obtain

$$\mathcal{L}_{l+1}(\mathcal{D}) \leq a\mathcal{L}_l(\mathcal{D}) + b \leq a^{l+1}\mathcal{L}_0(\mathcal{D}) + b\frac{1-a^{l+1}}{1-a} \quad (33)$$

by the pre-setting $0 < a < 1$, we obtain the main theorem.

**Remark 2** *Theorem 3 establishes the global convergence property of the estimation error in terms of a sum of spectral norms, and further implies that the value of $\mathcal{L}_l(\mathcal{D})$ is decaying exponentially after $l$ stages, and the algorithm is less sensitive to the initial values.*

## Sparse Neural Networks

The convolutional neural networks (CNN) are powerful models for learning visual patterns from raw images.Generally speaking, the practical networks have too many parameters to fit in mobile devices or on-chip. The pursuit of neural networks with fewer parameters draw a lot of attentions, dating back to many years ago (LeCun, Denker, and Solla 1990)(Hassibi, Stork, and others 1993). Although recent advances showed that with a trained network as a warm start, a network can be highly compressed. But their experiments are mostly done on networks with large FC layers and many redundant parameters. We will explore the potential of compressing model weights by directly introducing the sparse regularizer to the loss functions, with a small and compact network, without any warm start, purely optimizing a regularized function.

For a CNN, denoting an input features map of height $h$, width $w$, and channels $c_i$, as $I \in R^{h \times w \times c_i}$, the convolutional kernel $K$ is in shape size of $s$ and with $c_o$ output channels, then $K \in R^{s \times s \times m \times c_o}$.

$$O(y, x, j) = \sigma[\sum_i^{c_i} \sum_u^s \sum_v^s K(u, v, i, j) I(y+u-1, x+v-1)]$$

where $\sigma$ is a nonlinear function like ReLU. We use $y$ to represent the labels, and denote $K$ as the convolution kernels, then the objective function is written as

$$\min_K \frac{1}{N} \sum_{n=1}^N L(\mathcal{F}(I_n, K), y_n) + \sum H(K). \quad (34)$$

where $L$ is the loss function, $\mathcal{F}(I_n, K)$ is the output of network parameterized by $K$, and $H(K)$ is the LCNR mentioned earlier. Although the objective is mostly non-smooth due to the ReLU layers and max-pooling layers, we can calculate the proximal steps based on the subgradient, which is obtained by back-propagation. Since the weights across different layers are with significantly different statistics, we put separate regularizer on each layer, with independent threshold and coefficient, to induce sparsity of different degrees.

For neural networks, the data samples are much more than 3D recovery, for the multi-stage algorithm, the local optimal point can not be accessed assuredly, even with a convex regularizer. Therefore, we proposed a doubly majorization-minimization, which is practical to implement. The algorithm is based on the multi-stage, gradually calibrating the regularizer into regular $\ell_1$ regularizers in each stage. While in each stage, the solver is not required to reach the local optima. The algorithm builds another (stochastic) upper bounding function $g(c, c^{l,t-1}, \mathcal{S}_t)$ for the stage-wise surrogate function $L(c) + H^l(c)$, based on the stage-wise initializer point $c^{l,0}$ and the given subset $\mathcal{S}_t$. For a smooth objective, $g(c, c^{l,t-1}, \mathcal{S}_t)$ can be set to be the following proximal function,

$$g(c, c^{l,t-1}, \mathcal{S}_t) = \frac{1}{|\mathcal{S}_t|} \sum_{n \in \mathcal{S}_t} [L(x_n, c^{l,t-1})$$
$$+ \nabla L(x_n, c^{l,t-1})^T(c - c^{l,t-1})] + \frac{1}{2\eta}||c - c^{l,t-1}||^2.$$

where $\mathcal{S}_t$ is the selected subset of dataset in the $t$-th step, and $1/\eta$ can be set to the smoothness parameter (if exists). Minimizing the above function can be attained by taking a proximal gradient step, $c^{l,t} \leftarrow prox_H(c^{l,t-1} - \eta\nabla L(x_n, c^{l,t-1}))$.

---

**Algorithm 1** (Stochastic) Doubly MM Algorithms

  **Input** $\{x_n\}_{n=1}^N$. Initialize $\lambda_i^0 = \beta$, $l = 0$, $H^0 = \beta|c|$.
  **repeat**
    Update $\lambda_i^{l+1}$ as Eq.(4), and set $l = l+1$.
    **repeat** Solve $\min_c g(c, c^{l,t-1}, \mathcal{S}_t) + H^l(c)$ on minibatch $\mathcal{S}_k$ using proximal gradient descent.
  **until** Converge

---

## Experiments

### Sparse Linear Regression

We implement a sparse linear regression model, for testing the effect of LCNR in inducing sparsity. The dataset is constructed as following, matrix $X \in \mathbb{R}^{D \times N}$ stacks $N = 1000$ random variables of $D = 256$ dimensions, drew from a Gaussian distribution, most elements of ground truth weights $\bar{w} \in \mathbb{R}^D$ are zero, the others are drew from the Gaussian distribution, the target vector $Y \in \mathbb{R}^{1 \times N}$ is obtained by $Y = \bar{w}^T X$ plus a Gaussian noise $\delta$. The objective function is to minimize the $\frac{1}{2}||Y - w^T X||_F^2 + H(w)$, where we set $H(w)$ to leaky capped $\ell_1$ norm, or regular $\ell_1$ norm as a baseline. We use the multi-stage algorithm, which repeatedly takes gradient descent step and a proximal step. We set the maximal inner-iterations to be 20 per stage, and 50 stages in total as it converges well. We searched the regularization parameters and stepsize that perform best for problem, while not introducing considerable estimation bias, then, we also test several near parameters by $\times 1e-1$. For the LCNR, we set $\alpha/\beta = 300$.

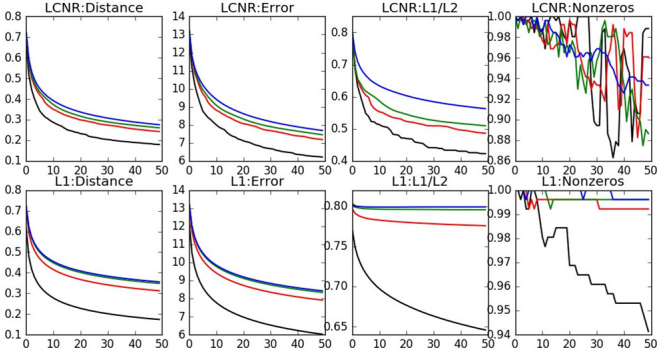

Figure 2: Performance under different measure. Black: $\beta = 100.0$, red: $\beta = 10.0$, green: $\beta = 1.0$, blue: $\beta = 0.1$. (1) the **distance** from current estimation to the ground truth as $||w^t - \bar{w}||_F$, (2) the **error** of linear regression as $||Y - X^T w^t||_F$, (3) the ratio of $\ell_1/\ell_2$ norm, as $||w||_1/||w||_2$, (4) the **nonzero** rate as $\mathbb{E}[\mathbb{I}(w_i^t \neq 0)]$.

We denote $w^t$ as the estimated weights in $t$-th stage. We measure the performance by four standards. The statistic numbers change against every stage is plotted in Figure.(2). As we see from the figure, the proposed model with LCNR converges to the ground truth with faster speed, achieving higher accuracy, and yields much sparser solutions than the $\ell_1$ norm.

**Human 3D Shape Recovery**

We conducted the expriments to verify the effectiveness of leaky capped $\ell_1$ regularization in learning sparse weights for 3D shape recovery. Our proposed algorithm is implemented in MATLAB, based on the code generously provided by (Zhou et al. 2016). We use the CMU motion capture dataset (moc ) for both training and testing, thousands of frames of 3D human shapes are contained within the dataset. The shapes are annotated by 3D locations of 15 landmarks, as $S \in \mathbb{R}^{3 \times 15}$, and landmarks are at anatomical joints of human, like head, shoulders, elbows, hips, ankles and etc. The dataset contains various kinds of action. As there are large external variations across actions, we take each single action into analysis, but using the same shape dictionary. We use 300 frames of each action as test set. The rest of frames are used as training set for building shape dictionary. We set $D = 128$ to construct an over-complete dictionary by common sparse coding algorithm with $\ell_1$ regularization, and the training data are pre-aligned by the Procrustes method used in (Ramakrishna, Kanade, and Sheikh 2012). The 2D shapes for test set are synthesized from the ground truth 3D shapes at different angles across 360 degrees. The recovery error is measured by the Frobenius norm $||\hat{S} - S||_F$ from the recovered 3D shape $\hat{S}$ with the ground truth $S$. We compare our method to state-of-the-art algorithm (Zhou et al. 2016) (which has been extensive compared against other methods in their paper, like Projected Matching Pursuit (Ramakr-ishna, Kanade, and Sheikh 2012) and the alternating manifold minimization method, and proved superiority, therefore we do not compare others). We conducted experiments comparing the LCNR with other representative (non)convex regularizations, including $ell_1$ norm, capped $ell_1$ norm, logarithm norm, and Laplace norm, as

$$\text{(Convex)} \quad \ell_1 : R(c; \lambda) = \lambda|c|,$$
$$\text{capped-}\ell_1 : R(c; \lambda, \tau) = \lambda \min(|c|, \tau),$$
$$\text{logarithm} : R(c; \gamma, \lambda) = \frac{\lambda}{\log(\gamma + 1)} \log(\gamma|c| + 1),$$
$$\text{Laplace} : R(c; \lambda, \gamma) = \lambda \left(1 - \exp(-\frac{|c|}{\gamma})\right)$$

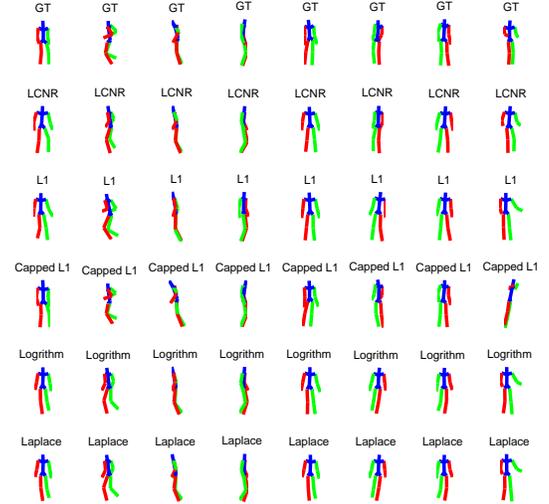

Figure 3: Recovery results of different kinds of actions, by using LCNR and other (non)convex regularizations, and the ground truth shape (GT) on the top row.

We continue to use the doubly MM algorithm to solve the models, and the constraint on dictionary coefficients $c$ is transferred to the spectral norm of 3D shapes $\{M_i\}$. The regularization parameters are grid searched for the best final performance. The computation complexity of calibrating $\tau$ and $\lambda_i^l$ is considerably smaller than the ADMM parts, the average running time of this part is about $14\%$ of the overall time of each iteration.

In Figure.(3), we plot the recovered shapes by LCNR regularized and $\ell_1$ regularized models, and the ground truth shapes, within a maximum inference iterations of 200. To test the performance against large noise, we also add matrix $[\sigma * mean(abs(S)) * randn(size(S))]$ to each 3D shape $S$ before generating the 2D observation $Y$. Due to the limited space, we put the results in appendix. The mean errors for the testing frames of different action types decreasing against stages are shown in Figure.(4). One can see the proposed model reconstruct much more accurate skeletons than state-of-the-art model. By comparing the convergence rate in Figure (4), we see that, the non-convex regularizations generally converge faster than the $\ell_1$ because they intro-duce less estimation bias, so a larger coefficient $\lambda$ can be adopted for acceleration. The capped norm induces faster convergence rate within the beginning 30 iterations, however, in the following iterations the estimation error oscillates up and down, indicating an unstable behavior, this indicates that a very sparse solution doesnot benefit the 3D recovery problem. The $\ell_1$ regularization can achieve good estimation at last, but in the beginning the error decreases very slow, due to the imbalance between accuracy and speed, but it outperforms capped $\ell_1$, indicating that a certain degree of regularization is critical to larger weights. We can see for most of the actions, LCNR leads to higher recovery accuracy comparing opponents, achieving the same accuracy requirement within much lesser time, this improvement is significant especially to test phase, which is most the case since training phase only needs accomplished once.

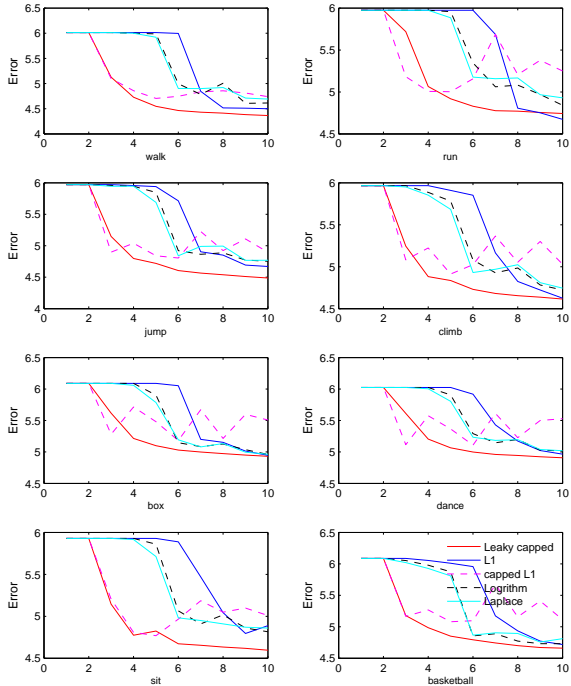

Figure 4: Convergence of recovery error with different regularizations. (X-axis: stages (for 10 inner iterations)

**Sparse Neural Networks**

We implement a convolutional neural networks, with 5 convolutional layers of $5 \times 5 \times 16$ and $3 \times 3 \times 32$, three pooling layers of $(2, 2)$, and a softmax loss layer. We use the CIFAR10 dataset. We compare the proposed LCNR with regular $\ell_1$ norm regularizer. The training mini-batch size is set to be 128 per iteration, and set the maximum data passes to be 40 rounds. During each data pass, we calculate the nonzero rate of model weights, the pure loss function value without the regularizers, and test the accuracy using the testing set. For the $\ell_1$ regularizer, we set parameter to be $\beta = \lambda_0$, and the parameter of LCNR is set according to $\alpha = \lambda_0 * 2.0$ and $\beta = \lambda_0 * 0.1$, therefore the penalty on larger weights are sig-nificantly lighter than smaller ones. We grid search $\lambda_0$ within $10.0 * 2^i$ for $i$ in range from $-8$ to $0$, obtaining the neural networks with different sparsity. We use the Adam solver for optimization. The initial learning rate is set to be $0.02$. After each iteration, we additionally perform the proximal gradient steps $x^{l,t} = prox_{\eta R}(x^{l,t-1} - \eta \partial L(x^{l,t-1}))$, to ensure the sparsity of convolutional filters. The proximal gradient steps are like $prox_{\eta R}(x) = \text{sign}(x) \odot \max(|x| - \eta \lambda, 0)$, where $\odot$ is the element-wise product. We plot the loss function values and sparse rates changing over time in Figure.(5) and the statistics in the final data pass in Table.(1). We jointly show the results of two methods in similar sparse rate. We can see that comparing to the $\ell_1$ regularizer, the proposed LCNR achieves better sparsity, without damaging much on the loss function, and boosting the accuracy by about $2\%$, for achieving the same sparsity.

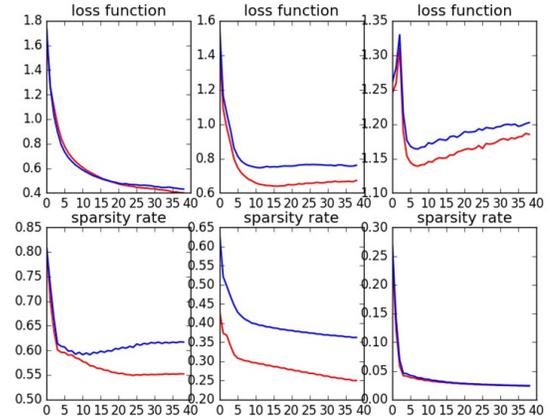

Figure 5: Statistics changing over data passes, and each column shows result of one setting. Red: LCNR regularized CNN, blue: $\ell_1$ regularized CNN.

Table 1: Performance Statistics (each column shows result of one setting).

| sparse level | method | accuracy | sparsity | loss |
|---|---|---|---|---|
| level-1 | LCNR | 70.9 | 55.2 | 0.40 |
|  | $\ell_1$ | 68.8 | 61.6 | 0.43 |
| level-2 | LCNR | 69.2 | 24.9 | 67.1 |
|  | $\ell_1$ | 68.3 | 36.2 | 76.0 |
| level-3 | LCNR | 55.8 | 2.3 | 1.18 |
|  | $\ell_1$ | 53.2 | 2.4 | 1.20 |

## Conclusion

In this paper, we present the leaky capped $\ell_1$ norm regularizer and its multi-stage optimizer. We apply it in monocular 3D shape recovery problem, obtaining an accuracy improvement and acceleration over state-of-the-art algorithm, and prove its global convergence, the first rigorous analysis in relevant literatures and non-convex problems. Auxiliary empirical studies over sparse linear regression and convolutional neural networks show the LCNR improves upon $\ell_1$ regularization.

# Appendix

To test the performance against large noise, we also add matrix $[\sigma * mean(abs(S)) * randn(size(S))]$ to each 3D shape $S$ before generating the 2D projection. Due to the limited space, we put the results in Figure 6;7;8;9.

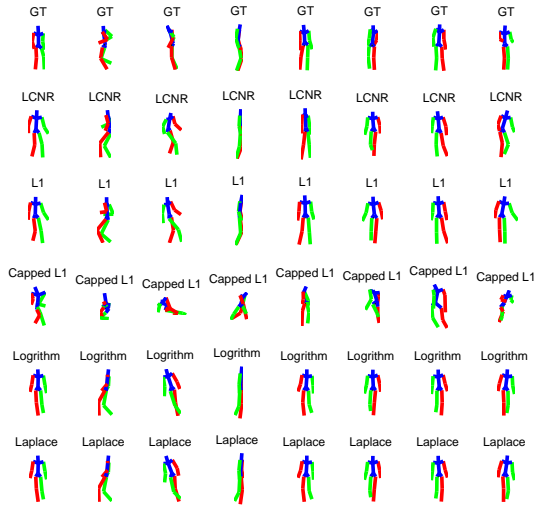

Figure 6: Recovery results of different kinds of actions, under noise condition $\sigma = 0.15$, by using LCNR and other (non)convex regularizations, and the ground truth shape (GT) on the top row.

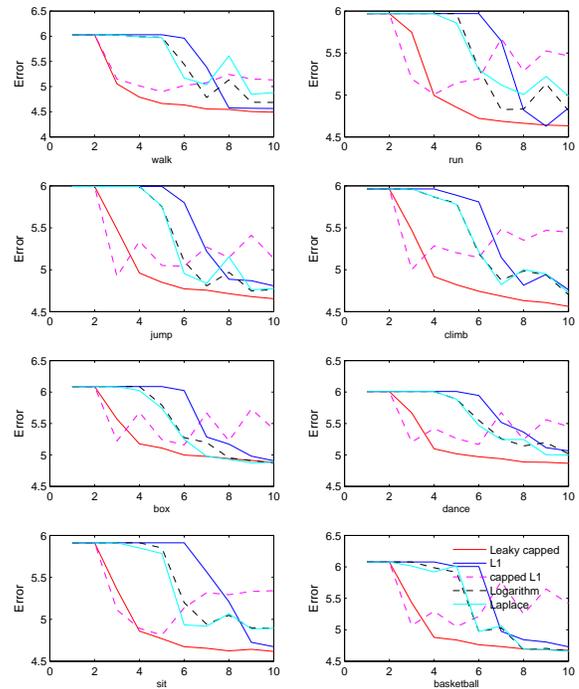

Figure 7: Convergence of recovery error under $\sigma = 0.05$ noise, with different regularizations. (X-axis: stages (for 10 inner iterations)

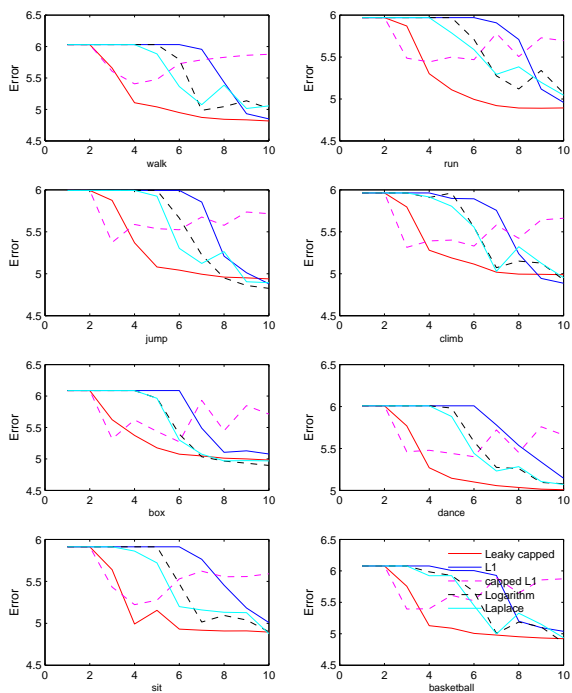
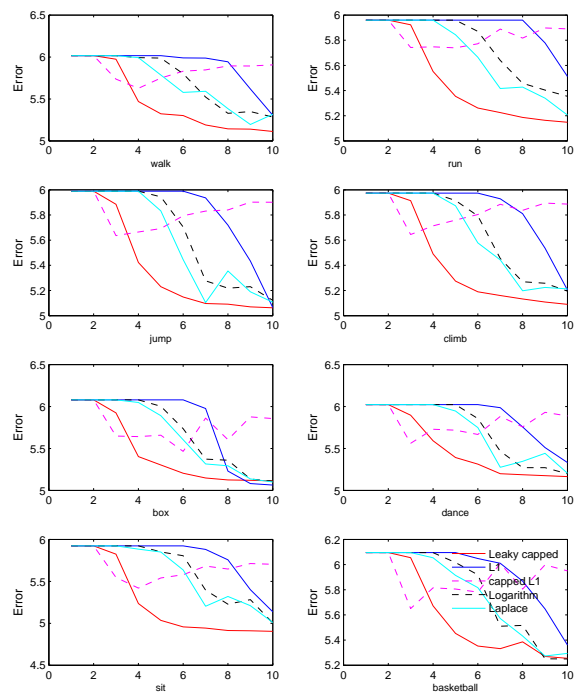

Figure 8: Convergence of recovery error under $\sigma = 0.15$ noise, with different regularizations. (X-axis: stages (for 10 inner iterations)

Figure 9: Convergence of recovery error under $\sigma = 0.25$ noise, with different regularizations. (X-axis: stages (for 10 inner iterations)